\pdfoutput=1

\documentclass[11pt]{article}

\usepackage{EACL2023}

\usepackage{times}
\usepackage{latexsym}
\usepackage{tikz}
\usepackage{amssymb}
\usepackage{amsmath}
\usepackage{lscape}
\usepackage{xcolor}
\usepackage{caption}
\usepackage{subcaption}
\usepackage{longtable}
\usepackage{natbib}
\usepackage{float}
\usepackage{graphicx}
\usepackage{booktabs}
\usepackage{multirow}
\usepackage{url}
\usepackage{soul}

\newcommand{\hlc}[2][yellow]{{%
    \colorlet{foo}{#1}%
    \sethlcolor{foo}\hl{#2}}%
}
\definecolor{brightgreen}{rgb}{0.4, 1.0, 0.0}

\usepackage[T1]{fontenc}

\usepackage[utf8]{inputenc}

\usepackage{microtype}

\usepackage{inconsolata}

\title{Implicit Temporal Reasoning for Evidence-Based Fact-Checking}

\author{Liesbeth Allein, Marlon Saelens, Ruben Cartuyvels, Marie-Francine Moens \\
        Department of Computer Science \\ KU Leuven, Belgium \\ \{liesbeth.allein,ruben.cartuyvels,sien.moens\}@kuleuven.be}

\begin{document}
\maketitle
\begin{abstract}

Leveraging contextual knowledge has become standard practice in automated claim verification, yet the impact of temporal reasoning has been largely overlooked. Our study demonstrates that time positively influences the claim verification process of evidence-based fact-checking. The temporal aspects and relations between claims and evidence are first established through grounding on shared timelines, which are constructed using publication dates and time expressions extracted from their text. Temporal information is then provided to RNN-based and Transformer-based classifiers before or after claim and evidence encoding. Our time-aware fact-checking models surpass base models by up to 9\% Micro F1 (64.17\%) and 15\% Macro F1 (47.43\%) on the MultiFC dataset. They also outperform prior methods that explicitly model temporal relations between evidence. Our findings show that the presence of temporal information and the manner in which timelines are constructed greatly influence how fact-checking models determine the relevance and supporting or refuting character of evidence documents.\footnote{The code of this paper is publicly available: \url{https://github.com/Marlon668/VerificationClaimsWithTimeAttribution}.}

\end{abstract}

\section{Introduction}

\begin{figure}[ht]
    \centering
    \includegraphics[scale=0.43]{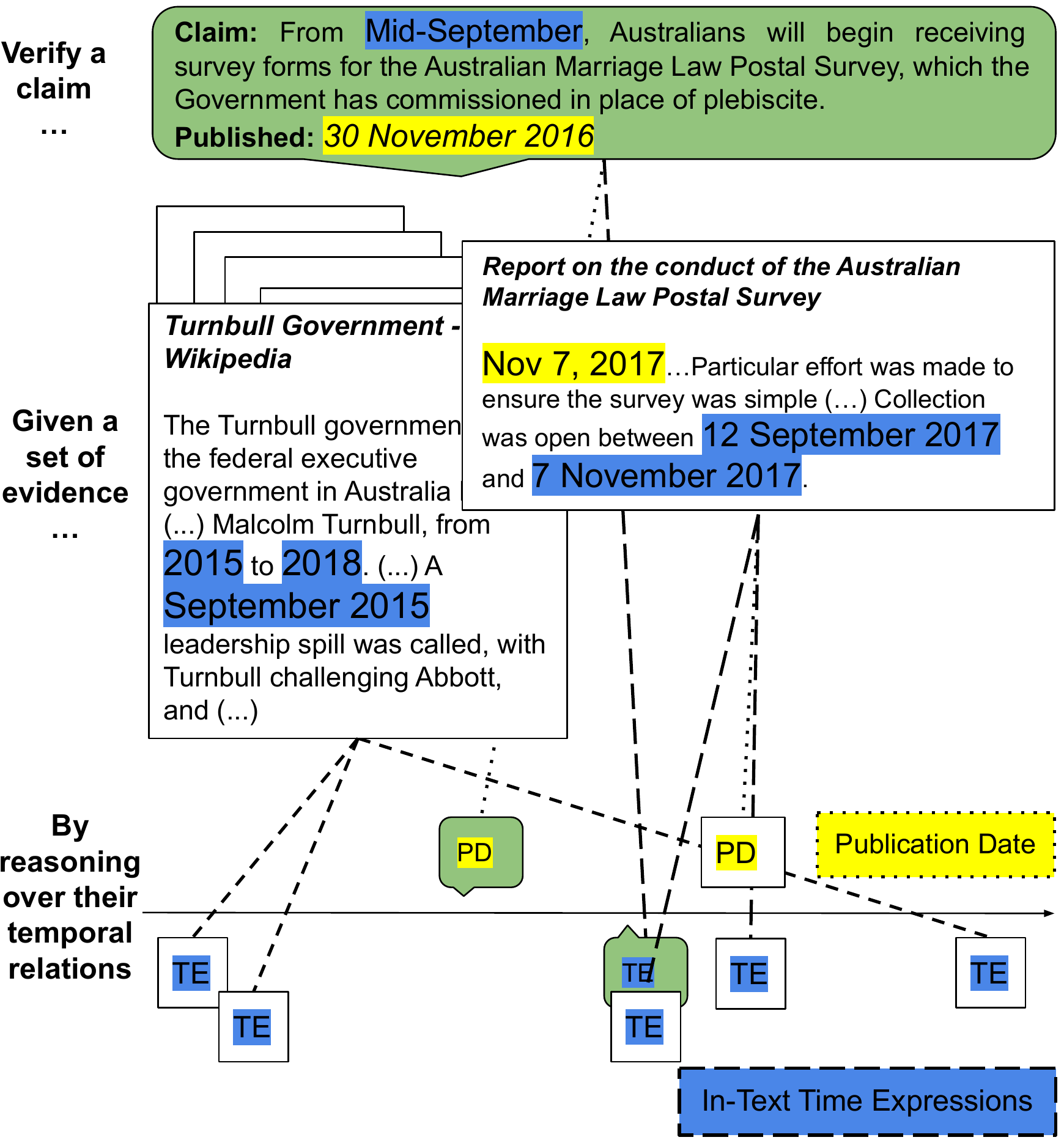}
    \caption{An evidence-based fact-checking model verifies a given claim against a set of Web documents serving as supporting or refuting evidence. In this study, we let the model implicitly reason over the temporal aspects of the claim and evidence, and their relations. For this, both inputs are grounded at two levels on a shared timeline: at the document level using their publication dates (in yellow, dotted line) and at the content-level using time expressions in their text (in blue, dashed line).}
    \label{fig:introduction}
\end{figure}

Automatically verifying information and flagging engineered falsities have been high on the political, media, and - subsequently - research agenda for quite some \textit{time} \citep{ECdisinfo}. However, the role of time in machine-assisted fact-checking has been inadequately investigated. Time can affect the veracity of previously uttered claims and the relevance of supporting or refuting evidence. This is evident in research, for example, where newly acquired knowledge may question, confirm, or refute established facts. This study proposes to ground claims and associated evidence in time and incorporate temporal reasoning abilities in the claim verification process of computational fact-checking models (Figure \ref{fig:introduction}). Here, temporal reasoning is implicit since the models are not expected to make explicit predictions about time. They instead learn from data how to leverage temporal information.

Grounding a claim or evidence document in time is a complex task. 
On the one hand, it can be achieved through document-level grounding, which involves positioning the entire document on a timeline based on its publication date.
On the other hand, a document may discuss several events that have occurred in the past, present, or future. To facilitate more fine-grained grounding on the content level, time expressions in the text are used to to place the document on multiple positions on a timeline. Such expressions can be explicit (e.g., 27 June 2022), implicit (e.g., Christmas 2022), and relative (e.g., mid-September), which may require additional temporal information for grounding \citep{strotgen2013multilingual,leeuwenberg2019survey}.
In this study, we ground claims and evidence on both the document and content level. This is accomplished by extracting and normalising their publication date and in-text time expressions, and subsequently relating them in terms of distance in time. This enables fact-checking models to reason over the temporal relations between a claim and its evidence on more than one level.

\paragraph{Contributions} This study demonstrates that %
reasoning over temporal aspects and relations of claims and evidence not only improves fact-checking models' prediction performance but also influences their estimation of the relevance and the supporting/refuting character of the evidence. The effects on performance are even reinforced when claims and evidence are grounded at both the document and content level, showing the appropriateness of multi-level temporal reasoning in automated fact-checking.

\section{Related Work}

Automated fact-checking is usually a two-phase process consisting of claim detection/selection and claim verification \citep{zeng2021automated,guo-etal-2022-survey}. Time is arguably important in both phases. When detecting and ultimately selecting claims to fact-check, fact-checkers heed the current interest of the public in certain topics and election cycles, and rank the claims accordingly \citep{allein2020checkworthiness}. Moreover, many selected claims mention dates or time periods \citep{hidey-etal-2020-deseption}. \citet{shaar-etal-2020-known} looked in the past and filtered out claims that are semantically similar to previously fact-checked claims to expedite the claim selection process.

While evidence-based claim verification has been widely studied \cite{zhong-etal-2020-reasoning,liu-etal-2020-fine,chen2021entity,si-etal-2021-topic,jin2022towards,xu2022evidence,hu-etal-2022-dual}, few studies explicitly focused on incorporating temporal reasoning in the verification process. \citet{zhou-etal-2020-probabilistic} constructed (entity, value, time)-tuples representing supposedly temporal facts and verified their correctness using probabilistic graphical models. \citet{allein2021time} constrained the evidence ranking in fact-checking models on time using silver-standard evidence rankings respecting four assumptions on temporal relevance. Instead of verifying the temporal correctness of claim tuples or explicitly enforcing time-dependent evidence rankings, we let fact-checking models reason \textit{implicitly} over temporal aspects of claims and evidence in natural language when checking the claims.

\section{Task Description} \label{taskdescription}

Classifier $f$ takes a textual claim $c$ and an associated set of $N$ text documents $\{e_i\}^N$ serving as evidence of $c$, and returns a claim veracity label $y$. 

\begin{equation}
    f: c, \{e_i\}^N \xrightarrow{} y 
\end{equation}

To allow $f$ to reason over temporal aspects of $c$ and $e_i$, we extract and normalise publication dates and time expressions in $c$ and $e_i$, and assign them to time buckets.
Temporal representations $c_t$ and $e_{i,t}$ are sequences of time bucket indices and are given as additional input to $f$:

\begin{equation}
    f: c, c_t, \{e_i\}^N, \{e_{i,t}\}^N \xrightarrow{} y  
\end{equation}

\begin{figure*}[ht]
    \centering
    \begin{subfigure}[b]{0.47\textwidth}
         \centering
         \includegraphics[width=\textwidth]{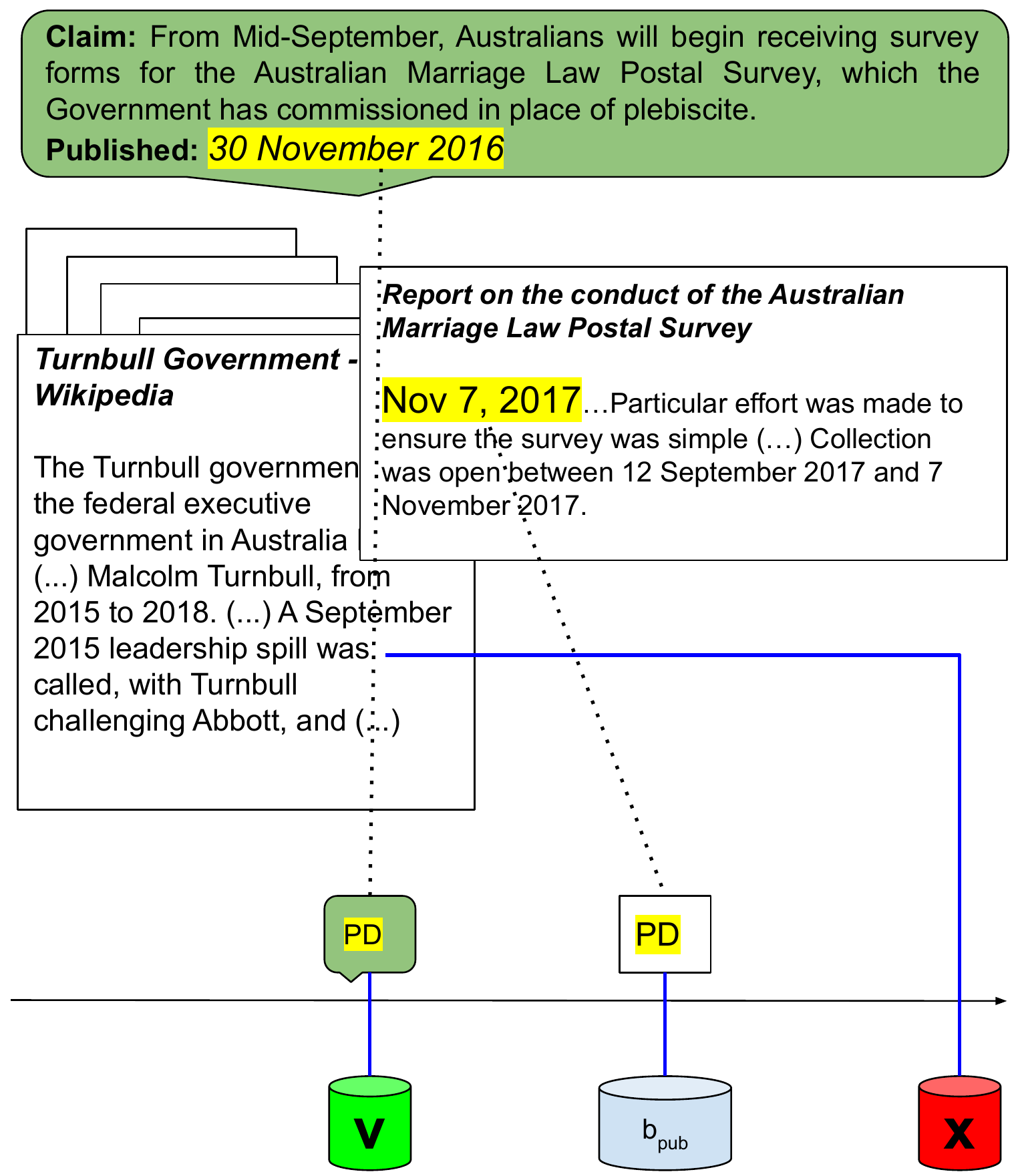}
         \caption{Document-Level Grounding}
         \label{fig:ex1}
     \end{subfigure}
    \begin{subfigure}[b]{0.47\textwidth}
         \centering
         \includegraphics[width=\textwidth]{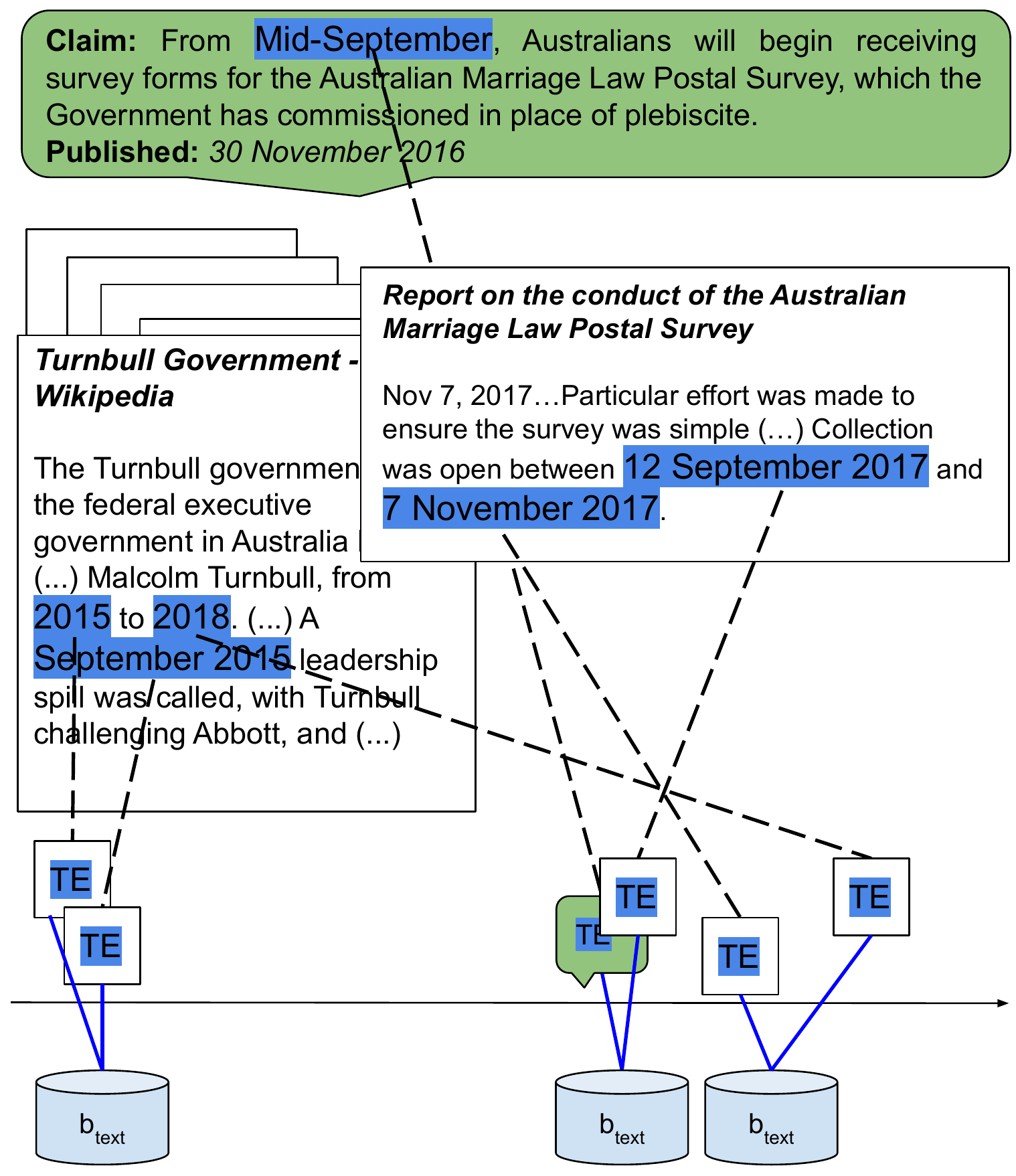}
         \caption{Content-Level Grounding}
         \label{fig:ex2}
     \end{subfigure}
    \caption{Illustration of two-level grounding: (a) at the document level using publication dates (\hlc[yellow!70]{PD}) and (b) at the content level using in-text time expressions (\hlc[blue!40]{TE}). 
    All PD and TE are assigned to time buckets $b_{pub}$ and $b_{text}$, respectively. 
    \hlc[brightgreen!80]{\textbf{V}} means that a publication date was found (only for claims) and \hlc[red!70]{\textbf{X}} that no publication date was found. 
    }
    \label{fig:groundingoverview}
\end{figure*}

\section{Two-Level Grounding and Reasoning}

To obtain temporal representations $c_t$ and $e_{i,t}$, we ground $c$ and $e_i$ in time by positioning them on a joint timeline using either their \textit{publication date} ($c_t = c^{doc}_t$; $e_{i,t} = e^{doc}_{i,t}$) or \textit{in-text time expressions} ($c_t = c^{con}_t$; $e_{i,t} = e^{con}_{i,t}$). A fact-checking model can then reason over their temporal aspects and relations at the \textit{document level} or the \textit{content level}, respectively (Figure \ref{fig:groundingoverview}).

\subsection{Reasoning at Document Level}\label{groundingdocument}

The publication date of $c$ serves as reference point for grounding $e_i$. This way, we lay bare the temporal relation between $c$ and $e_i$ at the document level. We adopt the approach of \citet{allein2021time} and compute the distance in days $\Delta pub\in \mathbb{Z}$ between the publication date of $c$ and that of $e_{i}$, where $\Delta pub < 0$ indicates that $e_i$ was published before $c$, $\Delta pub = 0$ indicates that $e_i$ and $c$ were published on the same day, and $\Delta pub > 0$ indicates that $e_i$ was published after $c$. 
The publication date of $e_i$ is then assigned to a time bucket $b_{pub} \in T^{doc}$ given $\Delta pub$.
Ultimately, the document-level temporal representation of $e_i$, $e^{doc}_{i,t}$, is a sequence of indices corresponding to $b_{pub}$ in $T^{doc}$, with $|e^{doc}_{i,t}| = 1$ since $e_i$ has only one publication date. 
When a publication date for $e_{i}$ could not be extracted, $e^{doc}_{i,t}$ corresponds to the index of a dedicated time bucket indicating date unavailability. 
Lastly, the document-level temporal representation of the claim, $c^{doc}_t$, merely indicates the availability of a publication date for the claim.
We motivate and discuss the choice of $T^{doc}$ in Section \ref{timebuckets}.

\subsection{Reasoning at Content Level} \label{groundingcontent}

While the document-level approach grounds $c$ and $e_i$ as whole documents, the content-level approach places them on various positions on a timeline using time expressions found in their text. Each time expression in $e_i$ and $c$ is first extracted and normalised, and its distance in days $\Delta exp \in \mathbb{Z}$ to the publication date of $c$ is computed. 
They are then assigned to time buckets $b_{text} \in T^{con}$ given $\Delta exp$. The choice of $T^{con}$ is discussed in Section \ref{timebuckets}.
The content-level temporal representation of $e_i$ is $e^{con}_{i,t}$ is a sequence of indices where each index corresponds to %
a $b_{text} \in T^{con}$. The length of $e^{con}_{i,t}$ equals the number of time expressions found in $e_i$, and the $j$\textsuperscript{th} element of $e^{con}_{i,t}$ corresponds to the index of the time bucket of the $j$\textsuperscript{th} time expression in $e_i$. A time bucket index can occur multiple times in $e^{con}_{i,t}$. The same grounding procedure is applied to obtain content-level temporal representation $c^{con}_{t}$ for $c$.
In contrast to $c^{doc}_t$, $c^{con}_t$ does not merely reflect the availability of a publication date for $c$ but grounds time expressions in the claim text with respect to the claim's own publication date.
The content-level grounding approach allows a fact-checking model to reason over the temporal aspects of the events discussed in $e_i$ and $c$, and their temporal relation to the publication date of $c$.

\subsection{Creating Time Buckets}\label{timebuckets}

Time buckets $b_{pub} \in T^{doc}$ and $b_{text} \in T^{con}$ represent time intervals with respect to the publication date of $c$ (e.g., $b_{pub} =$ [1, 4] indicates that $e_i$ was published between 1 and 4 days after $c$ had been published). Following the cluster hypothesis of \citet{jardine1971use} which states that documents in a cluster contain similar information, the similar information in a bucket is the distance in time to $c$.  
For document-level grounding and reasoning, the construction and choice of $T^{doc}$ goes as follows: (1) $\Delta pub$ for each $e_i$ in the training set is computed; (2) all $\Delta pub$ are ordered in ascending order; (3) and, finally, all $\Delta pub$ are subdivided in 20 quantiles, containing a similar number of $e_i$ ($\mu=8530.5, \sigma=266.87$). Each quantile represents one bucket $b_{pub}$. %
Various numbers of quantiles were tested, and 20 returned the best performance on the validation set. Three buckets denoting a lacking publication date for $e_i$, an available publication for $c$, and a lacking publication date for $c$ are added; hence, $|T^{doc}| = 23$.
A similar procedure is applied for constructing $T^{con}$ using $\Delta exp$ ($|T^{con}|=24$, $\mu=13390.75$, $\sigma=2050.4$). However, no extra buckets $b_{text}$ denoting (un)availability of date are added.
An overview of all $b_{pub}$ and $b_{text}$ can be found in Appendix \ref{app:timebuckets}. Note that the intervals of $b_{pub}$ and $b_{text}$ become smaller when its bounds approach 0, allowing for more fine-grained reasoning for evidence published around or at the same time as the claim. Time buckets approaching 0 (i.e., $e_i$ situates around the same time as $c$) have smaller intervals than those far from 0, with even a dedicated time bucket for those evidence published or discussing events happening on the same day as the claim. The advantage of using such time buckets 
is that the model is more robust against bias towards larger buckets. In the fact-checking models, each bucket corresponds to a unique embedding stored in a randomly-initialised time embedding matrix, which is updated during model training. 

\section{Methodology}

\subsection{Fact-Checking Model}

\begin{figure}[ht]
    \centering
    \includegraphics[scale=0.525]{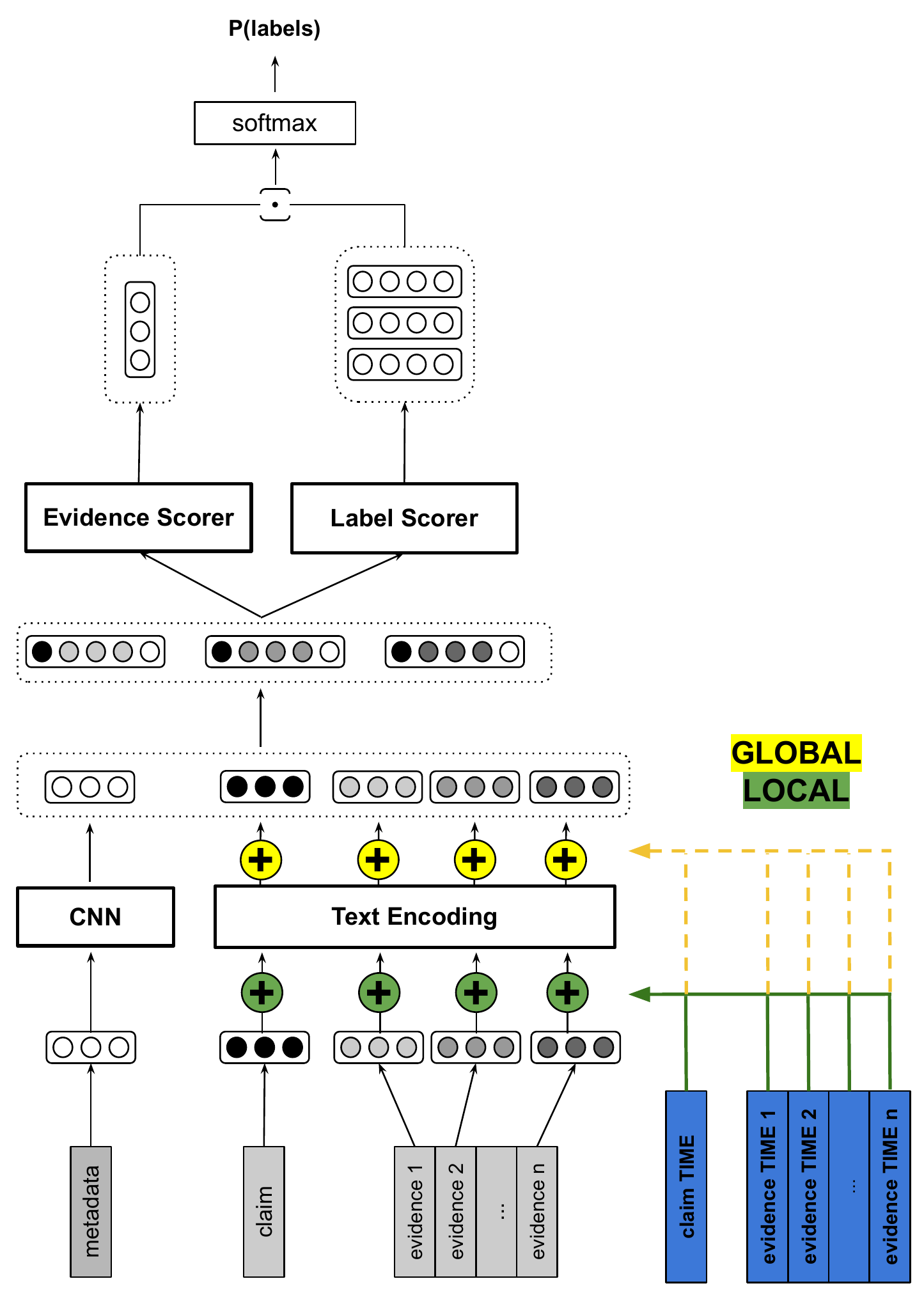}
    \caption{Overview of the fact-checking model, where temporal information on claim and evidence (in blue) is integrated before text encoding (local level; in green) or after text encoding (global level; in yellow).}
    \label{fig:model}
\end{figure}

We take the Joint Veracity Prediction and Evidence Ranking model introduced in \citet{augenstein-etal-2019-multifc} as base model (Figure \ref{fig:model}). 
Taking $c$ and $e_i$ represented by their word embeddings $w \in \mathbb{R}^{D_1}$, the text encoder encodes them to their latent representations $h(c)$ and $h(e_i)$ $\in \mathbb{R}^{D_2}$. Metadata $m$ linked to $c$ is encoded in parallel, yielding $g(m)$. Next, $h(c)$, $h(e_i)$, and $g(m)$ are combined into a joint claim-evidence representation $s_i$ using the matching approach introduced by \citet{mou-etal-2016-natural}:
\begin{equation}
    s_i = [h(c);h(e_i);h(c)-h(e_i);h(c) \cdot h(e_i);g(m)]
\end{equation}
with $[;]$ denoting concatenation, and $[\cdot]$ the dot product. 
The evidence scorer projects each $s_i$ to $o_i\in \mathbb{R}$, forming evidence score vector $o\in \mathbb{R}^{N}$. The label scorer projects each $s_i$ to its label score vector $q_i\in \mathbb{R}^{L}$ forming scoring matrix $Q\in \mathbb{R}^{N\times L}$, with $L$ the number of veracity labels. $o^\intercal \cdot Q$ gives a final score vector for all labels $L$, to which a softmax is applied to obtain a probability distribution over all veracity labels.

\subsection{Incorporating Temporal Reasoning} \label{incorporatingtime}

Temporal representations $c_t$ and $e_{i,t}$ are transformed to their time embeddings, $\hat{c}_t$ and $\hat{e}_{i,t}$, and given as additional model input. %
The embedding dimensions depend on the stage at which they are integrated in the model. 

\paragraph{Local integration} 
$c_t$ and $e_{i,t}$ are integrated before encoding claim and evidence $c$ and $e_i$ to $h(c)$ and $h(e_i)$. 
Time embeddings $\hat{c}_t, \hat{e}_{i,t}$ for 
each time bucket index in $c_t$ and $e_{i,t}$ are taken from the embedding matrix and projected onto the same dimension as the word embeddings $w \in \mathbb{R}^{D_1}$ of tokens in $c$ and $e_i$ using a linear transformation layer $l$. 

For \textbf{document-level reasoning} (DL$_{loc}$, eq. \ref{dl_loc}), the embeddings (there is max. one publication date; hence one time embedding per document) are prepended to those of $c$ and $e_i$. These are then sent to the text encoder.

\begin{equation}\label{dl_loc}
\begin{split}
    c &= [l(\hat{c}^{doc}_t); w_0, ..., w_{|c|}] \\
    e_i &= [l(\hat{e}^{doc}_{i,t}); w_0, ..., w_{|e_i|}] 
\end{split}
\end{equation}

For \textbf{content-level reasoning} (CL$_{loc}$, eq. \ref{clloc}), local integration is more complex. Firstly, $c^{con}_t$ and $e^{con}_{i,t}$ may refer to more than one time bucket as there may be more than one time expression in $c$ and $e_i$.
Secondly, the position of a time expression and the predicate it belongs to may provide rich information about a mentioned event. 
We first identify the type of each token in $c$ and $e_i$ (see Table \ref{tab:examplelocal}). 

\begin{table}[!h]
\centering
\resizebox{\columnwidth}{!}{
\begin{tabular}{|c|c|c|c|c|c|c|c|c|c|}
\hline
\multicolumn{10}{|l|}{\textbf{Position, predicate, and time expression marking}} \\
\hline
\textbf{Tokens} & Storm & Al- & berto & expected & to & make & landfall & to- & morrow \\
\hline
\textbf{Type} & O&O&O&O&O&\textsc{b-pred} & O & \textsc{b-time} & \textsc{time}\\
\hline
\textbf{Pos} &/&/&/&/&/&+2&/&/&/ \\
\hline
\textbf{Time} &/&/&/&/&/&/&/&0&/ \\
\hline
\end{tabular}}
\caption{Additional sentence preprocessing when integrating $c^{con}_t$ and $e^{con}_{i,t}$ at the local level. The predicate (\textsc{pred}) and the time expressions (\textsc{time}) are marked, with \textsc{b} indicating their first token, and the token distance between \textsc{b-time} and \textsc{b-pred} is computed.}
\label{tab:examplelocal}
\end{table}

We then introduce three new embeddings: predicate embedding $pr \in \mathbb{R}^{D1}$ marks the predicate, position embedding $po \in \mathbb{R}^{D1}$ marks the position of the predicate, and expression embedding $te \in \mathbb{R}^{D1}$ marks the time expression.
These additional embeddings are learned during training.
The word embedding $w$ of a token 
in $c$ depends on that token's type (same for $e_i$ and $e^{con}_{i,t}$):

\begin{equation}\label{clloc}
\displaystyle w = 
\begin{cases}
\displaystyle
\gamma w + (1-\gamma) (l(\hat{c}^{con}_{t,j}) + te) & \text{\scriptsize if \textsc{b-time}} \\ 
\displaystyle
\gamma w + (1-\gamma) te & \text{\scriptsize if \textsc{time}} \\
\displaystyle
\gamma w + (1-\gamma) (pr + po) & \text{\scriptsize if \textsc{b-pred}} \\ 
\displaystyle
\gamma w + (1-\gamma) pr & \text{\scriptsize if \textsc{pred}} \\
\displaystyle
w & \text{\scriptsize otherwise}
\end{cases}
\end{equation}

Embedding $\hat{c}^{con}_{t,j}$ refers to the embedding of time bucket $b_{text}$ to which the $j^\text{th}$ time expression in $c$ refers.

\paragraph{Global integration} $c_t$ and $e_{i,t}$ are integrated after $c$ and $e_i$ have been transformed by the text encoder to their latent representations $h(c)$ and $h(e_i)$ $\in \mathbb{R}^{D_2}$.  
An embedding for each time bucket in $c_t$ and $e_{i,t}$ is taken and projected onto the same embedding space $\mathbb{R}^{D_2}$ using a linear transformation layer $k$.
If $c_t$ or $e_{i,t}$ are represented by more than one time bucket, the embeddings are averaged. %
Fusion is performed using a weighted sum.
For \textbf{document-level reasoning} (DL$_{glob}$):
\begin{equation}\label{dl_glob}
\begin{split}
    h(c) &= \alpha h(c) + (1-\alpha) k(\hat{c}^{doc}_t) \\
    h(e_i) &= \alpha h(e_i) + (1-\alpha) k(\hat{e}^{doc}_{i,t})
\end{split}
\end{equation}
And for \textbf{content-level reasoning} (CL$_{glob}$):
\begin{equation}\label{cl_glob}
\begin{split}
    h(c) &= \alpha h(c) + (1-\alpha) \text{Avg}(k(\hat{c}^{con}_t)) \\
    h(e_i) &= \alpha h(e_i) + (1-\alpha) \text{Avg}(k(\hat{e}^{con}_{i,t}))
\end{split}
\end{equation}

with $\text{Avg}$ the average. We also experiment with a \textbf{combination of document-level and content-level reasoning} (DL+CL$_{glob}$, eq. \ref{dlcl_glob}) where temporal information from both levels is provided to the model:

\begin{equation}\label{dlcl_glob}
\begin{split}
    h(c) &= \alpha h(c) + \beta k(\hat{c}^{doc}_t) \\
     & + (1-\alpha-\beta) \text{Avg}(k(\hat{c}^{con}_t)) \\
    h(e_i) &= \alpha h(e_i) + \beta k(\hat{e}^{doc}_{i,t}) \\
     & + (1-\alpha-\beta) \text{Avg}(k(\hat{e}^{con}_{i,t}))
\end{split}
\end{equation}

\section{Experiments}

\subsection{Dataset}\label{dataset}

Experiments are conducted on MultiFC\footnote{The data is publicly available \href{https://competitions.codalab.org/competitions/21163}{on CodaLab}.} \citep{augenstein-etal-2019-multifc}, a large-scale dataset containing 34,924 English claims from various fact-checking websites (= `domains') where each claim is associated with at most 10 \textit{a posteriori} retrieved Web documents (319,721 documents in total). It also provides metadata on speaker, category, tags, and linked entities regarding the claim. We refer to \citet{augenstein-etal-2019-multifc} for a more detailed description of the data. Although other datasets for fact-checking have been proposed \citep{zeng2021automated}, they either lack naturally occurring claims, publication dates, or multiple evidence documents \citep{thorne-etal-2018-fever,jiang-etal-2020-hover,ostrowski2021multi,schuster-etal-2021-get}. Nonetheless, the large size, wide diversity of topic and data sources, and high quality of the MultiFC dataset should be sufficient for showcasing the appropriateness of our approach.

\subsection{Time Extraction and Normalisation} \label{extractionnormalisation}

In this section, we discuss the procedure for extracting and normalising publication dates and in-text time expressions.

\subsubsection{Publication Dates} 
The dataset provides the publication date of a claim as structured metadata. The date is represented as Year-Month-Day using rule-based temporal tagger HeidelTime \citep{strotgen2013multilingual}. The publication date of an evidence document, however, is not given in the metadata. Since its publication date is often communicated before the ellipsis (`...') at the beginning of its text,
we can automatically extract the date from the text \citep{allein2021time}. If we cannot extract a date at that position, we look for occurrences of \textit{`published'} or \textit{`posted'} in combination with a date. We again use HeidelTime for structuring the publication dates. In total, we obtain a publication date for 34,808 (99.67\%) claims and 213,165 (66.67\%) evidence documents.

\subsubsection{In-Text Time Expressions} Extracting and normalising in-text time expressions is more challenging as they can be implicit or relative. Since in-text time expressions are usually not annotated in datasets used for fact-checking, we need to reside to pretrained methods for extracting them. We implement the Open Information Extraction (OIE) model of \citet{stanovsky-etal-2018-supervised}, which parses a sentence and labels its arguments.
In this work, we focus on temporal arguments (${ArgM-TMP}$). Since inaccurate use or absence of capital letters has been shown to decrease the performance of OIE models \citep{alam2018domain}, the OIE model is expected to return a high number of inaccurate parses for capitalised news headlines -- which make up a large portion of the claims in the data. We therefore implement a pretrained Named Entity Recognition (NER) model \citep{peters-etal-2017-semi} to first detect people, locations, and organisations in the text. Then, the first token of each entity is capitalised while all other tokens are lowercased. Although capitalised temporal expressions such as weekdays and holidays are automatically lowercased too, we observed a higher quality of OIE parses when adopting this approach. We normalise the extracted temporal expressions using HeidelTime. The document creation time (DCT) of a piece of information, in this study the publication date, is used as reference point for normalising in-text temporal expressions. 
In total, we obtain 321,278 in-text time expressions.

\paragraph{Quality assessment} Implementing pretrained extraction and normalisation models inevitably introduces noise in the data. We therefore manually assess the quality of the NER, OIE, and HeidelTime models to ensure that the noise is limited. The assessment is performed on a randomly selected set of 10 claims and their accompanying evidence documents (104 in total) from the dataset, and performance is measured using precision (P), recall (R), and F1. Regarding NER, we investigate whether all entities have been recognised and completely extracted. The label correctness does not need to be evaluated. NER performance is 0.9054/0.9134/0.9094 (P/R/F1). For the OIE task, we assess whether all temporal expressions have been correctly extracted and parsed. OIE performance is 0.9608/0.5568/0.7050 (P/R/F1), indicating that while quite some time-related expressions have not been extracted, those found have been correctly parsed. Lastly, we evaluate the normalisation of the found expressions: HeidelTime performance is 0.9736/0.8409/0.9024 (P/R/F1). In all, we deem the quality of the pretrained extraction and normalisation models sufficiently high.

\subsection{Experimental Setup}

\paragraph{Hyperparameter settings} Both $c$ and $e_i$ are tokenised\footnote{Huggingface implementation of the DistilRoBERTa tokenizer: \href{https://huggingface.co/sentence-transformers/all-distilroberta-v1
}{sentence-transformers/all-distilroberta-v1}.} and represented using word embeddings (size $= 300$ (BiLSTM); $768$ (DistilRoBERTa)). We experiment with two neural text encoders for encoding $c$ and $e_i$: a two-layered bidirectional LSTM with skip-connections (dropout $= 0.1$, hidden size $= 128$) and a pretrained Sentence-DistilRoBERTa, which is a faster, distilled version of Sentence-RoBERTa \citep{sanh2019distilbert,reimers2019sentbert}. For sake of brevity, we continue to refer to this model as RoBERTa. Metadata $m$ is represented as a one-hot vector and encoded by a CNN (filter size $= 3$, kernel size $= 3$) with ReLU activation and 1D max pooling. The label scorer consists of two fully-connected layers (hidden size $= 100; 50$), both with ReLU activation. The evidence scorer is a fully-connected layer (hidden size $= 100$) with Leaky ReLU activation. All parameters except those of the pretrained RoBERTa model are initialised following a Xavier Uniform distribution. More detailed settings for reproducing the experiments, such as hyperparameter tuning, is provided in Appendix \ref{sec:reproducibility}.

\begin{table*}[]
    \centering
    \footnotesize
    \setlength\tabcolsep{4pt}
    \begin{tabular}{lllllll}
        \toprule
         & \multicolumn{3}{c}{\textbf{BiLSTM}} & \multicolumn{3}{c}{\textbf{RoBERTa}} \\
         & Micro F1 & Macro F1 & Fusion Weights & Micro F1 & Macro F1 & Fusion Weights \\
         \midrule
        Base & ${.5520}$ {\tiny ${(.0023)}$} & ${.3239}$ {\tiny ${(.0064)}$} & - & ${.6952}$ {\tiny ${(.0195)}$} & ${.5532}$ {\tiny ${(.0246)}$} & - \\
         \midrule
         DL$_{loc}$ & ${.5501}$ {\tiny $(.0095)$} & ${.3343}$  {\tiny $(.0277)$} & - & ${.5640}$ {\tiny $(.0084)$} & ${.3357}$ {\tiny $(.0174)$} & - \\
         DL$_{glob}$ & ${.6006}$ {\tiny ${(.0090)}$} & ${.4271}$ {\tiny ${(.0107)}$} & $\alpha=0.90$ & $\mathbf{.6973}$ {\tiny $(.0439)$} & ${.5608}$ {\tiny $(.0488)$} & $\alpha=0.75$ \\
         \midrule
         CL$_{loc}$ & ${.6098}$ {\tiny $(.0028)$} & ${.4491}$  {\tiny $(.0120)$} & $\gamma=0.50$ & ${.5685}$ {\tiny $(.0075)$} & ${.3601}$ {\tiny $(.0090)$} & $\gamma=0.10$ \\
         CL$_{glob}$ & ${.6089}$ {\tiny ${(.0167)}$} & ${.4425}$ {\tiny ${(.0167)}$} & $\alpha=0.25$ & ${.6882}$ {\tiny $(.0208)$} & $\mathbf{.5744}$ {\tiny $(.0376)$} & $\alpha=0.10$ \\
         \midrule
         DL+CL$_{glob}$ & $\mathbf{.6417}$ {\tiny ${(.0033)}$} & $\mathbf{.4743}$ {\tiny ${(.0080)}$} & $\alpha=0.20$ & ${.6947}$ {\tiny $(.0135)$} & ${.5739}$ {\tiny $(.0332)$} & $\alpha=0.20$ \\
          & & & $\beta=0.40$ & & & $\beta=0.20$ \\
         \bottomrule
    \end{tabular}
    \caption{Average test results over three (BiLSTM) and two (RoBERTa) runs - with standard deviation in brackets - aggregated over all 26 fact-checking domains. Experiments are conducted for document-level (DL) and content-level (CL) temporal reasoning, where temporal information is integrated before ($loc$) or after ($glob$) encoding.}
    \label{tab:results}
\end{table*}

\paragraph{Pretraining and fine-tuning} The experiments are conducted on the disjunct, label-stratified train (80\%), validation (10\%), and test set (10\%) provided by \citet{augenstein-etal-2019-multifc}. We adopt the pretraining and fine-tuning setup of \citet{allein2021time} to ensure transparent comparison. During pretraining, the model is trained on all 26 fact-checking domains where each domain is only presented once in each epoch (batch size = $32$ (BiLSTM); $16$ (RoBERTa)), mitigating model bias towards larger domains. After each epoch, the batch order is randomly shuffled,
and Adam with linear scheduler (lr = 1$e^{-4}$ (BiLSTM)) or RMSprop (lr = 2$e^{-4}$ (RoBERTa)) optimizes the model parameters using the cross-entropy loss on the prediction output. The best-performing model for each fact-checking domain is selected based on the validation loss. Each domain-specific model is then fine-tuned on only data from that domain and the best-performing model is again selected based on the validation loss.

\section{Results}

Table \ref{tab:results} reports model performance on the test set, aggregated over all domains, in terms of Micro F1 and Macro F1\footnote{Computed using the \href{https://scikit-learn.org/stable/modules/generated/sklearn.metrics.f1_score.html}{scikit-learn Python package}.}. The results show that the effect of temporal reasoning depends on (a) the level at which temporal information is integrated in the model (global vs. local), (b) the grounding/reasoning level (document vs. content), and (c) the model architecture (BiLSTM vs. RoBERTa). 
Regarding the integration level, global integration ($glob$) substantially surpasses local integration ($loc$) for document-level reasoning (both models; .5501/.3343 $\xrightarrow{}$ .6006/.4271 [BiLSTM]; .5640/.3357 $\xrightarrow{}$ .6973/.5608 [RoBERTa]) and content-level reasoning (.5685/.3601 $\xrightarrow{}$ .6882/.5744 [RoBERTa]).
Regarding the temporal grounding and reasoning level, the results show that the combination setup where claim and evidence are grounded at both the document and content level (DL+CL) yields the overall highest performance for BiLSTM (.6417/.4743), while marginally improving RoBERTa by 2\% Macro F1 (.5739). Lastly, temporal reasoning appears to impact the prediction performance of the less parameterised BiLSTM model more strongly than that of the Transformer-based RoBERTa model: .5520/.3239 $\xrightarrow{}$ .6417/.4743 [BiLSTM]; .6952/.5532 $\xrightarrow{}$ .6947/.5739 [RoBERTa]. A similar effect was observed by \citet{allein2021time}, who explicitly modeled temporal relations between a claim and its evidence by constraining model parameters on evidence rankings following various assumptions on temporal relevance. This could be attributed to the expressive power of large pretrained Transformers-based language models and the orders of magnitude of their pretraining set size. 

Table \ref{tab:results-baselines} shows the comparison between our best performing set-up with the baseline from \citet{augenstein-etal-2019-multifc} and the model with explicit temporal reasoning from \citet{allein2021time}. 
Overall, our approach outperforms the baseline and the explicit temporal reasoning approach, especially on the Macro F1-score.
This demonstrates the appropriateness of our implicit, two-level temporal reasoning method over an approach without temporal reasoning and one that explicitly models temporal relations using only publication dates.

\begin{table*}[]
    \small
    \centering
    \begin{tabular}{lllll}
        \toprule
         & \multicolumn{2}{c}{\textbf{BiLSTM}} & \multicolumn{2}{c}{\textbf{Transformer}}  \\
         & Micro F1 & Macro F1 & Micro F1 & Macro F1 \\
        \midrule
        No temporal reasoning \citep{augenstein-etal-2019-multifc} & .5520 & .3239 & .6952 & .5532  \\
        Explicit temporal reasoning \citep{allein2021time} & .6265 & .3673 & .5921$^\dagger$ & .3135$^\dagger$ \\ 
        \midrule
        Implicit temporal reasoning (Ours) & .6417 & .4743 & .6947 & .5739 \\
        \bottomrule
    \end{tabular}
    \caption{Results of our implicit temporal reasoning approach vs. the baseline results of \citet{augenstein-etal-2019-multifc} (our implementation) and the explicit temporal reasoning method of \citet{allein2021time}, with a BiLSTM and a Transformer text encoder. $^\dagger$: DistilBERT \citep{sanh2019distilbert} instead of RoBERTa.}
    \label{tab:results-baselines}
\end{table*}

\section{Discussion}

\paragraph{Weighting text and time} We ran experiments with various weight values ($\alpha, \beta, \gamma$) for combining the text features of a claim and its evidence with their temporal information\footnote{A full overview of tested values and the tuning approach is provided in Appendix \ref{app:timebuckets}.}. Table \ref{tab:results} presents the best-performing weight values for each setting based on the validation loss. 
When reasoning over the document-level temporal relations (DL), the results suggest that higher importance should be attributed to the text of the claim and its evidence rather than to their temporal information. However, this is the opposite when reasoning at the content level (CL). The combined setup (DL+CL) aligns with (CL) by attributing more importance to time than text. This suggests that specially in-text time expressions carry useful information for fact-checking a claim.

\paragraph{Impact on evidence relevance and label scores}

We analyse how and to which extent temporal reasoning influences a model's assessment of the relevance ($o_i$) and supporting/refuting nature ($q_i$) of evidence in relation to a claim. Since the model computes $o_i$ and $q_i$ for each evidence document associated with the claim, a ranking of all evidence can be derived based on either $o_i$ or $q_i$. We then measure the difference in such rankings between the base and the best-performing temporal models. 
Following
\citet{allein2021time}, we rely on the Spearman's rank correlation $r_s$, which is a non-parametric, distribution-independent metric for computing the correlation between two rankings. The correlation between the base model and the temporal reasoning models with regard to evidence relevance ranking is very weak, with $0<|r_s|<0.19$ for both BiLSTM and RoBERTa. Also between the temporal models, those correlations are generally very weak. 
Interestingly, the impact of implicit temporal reasoning on a fact-checking model's estimation of evidence relevance is arguably as strong as when performing explicit temporal reasoning \citep{allein2021time}. The correlations fall within the range of .17 < $|r_s|$ < .24. 
The correlations regarding label scoring ($q_i$) are comparable to those for evidence ranking, ranging from weak ($0.2<|r_s|<0.39$) and to very weak.
We can thus conclude that a model's estimation of the relevance and supporting/refuting nature of evidence documents is strongly influenced not only by the ability to reason over time, but also by the way a claim and its evidence are grounded on a timeline.

\begin{figure}[t]
\centering
  \begin{subfigure}[b]{\columnwidth}
  \includegraphics[trim=0.6cm 0.6cm 0.6cm 0.6cm,clip,width=\textwidth]{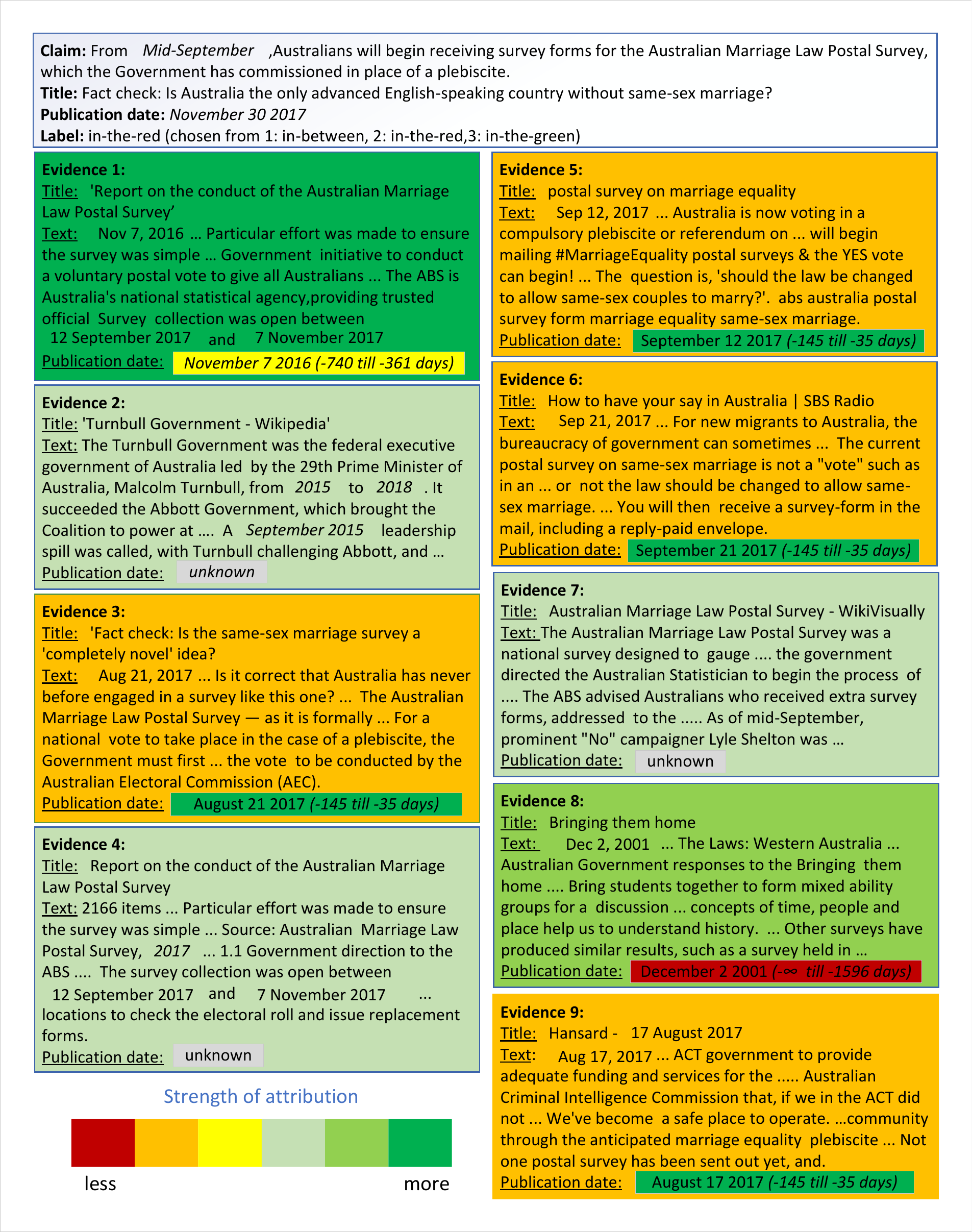} 
  \caption{Ranking of evidence by attribution strength in terms of text and publication date (DL reasoning).}
  \label{subfig:ranking}
  \end{subfigure}
  \begin{subfigure}[b]{\columnwidth}
  \centering
  \footnotesize
  \resizebox{\linewidth}{!}{
    \begin{tabular}{ccccc}
        \toprule
         & Base & DL & CL & DL+CL \\
        \midrule
        Label & (1) {\small5.3$e^{-4}$} & (1) {\small1.1$e^{-7}$} &(1) {\small.076} & (1) {\small.174} \\
        distribution & (2) {\small2.5$e^{-3}$} & \textbf{(2)} {\small\textbf{.530}}(\checkmark) & (2) {\small.172} & (2) {\small.266} \\
         & \textbf{(3)} {\small\textbf{.996}} & (3) {\small.470} & \textbf{(3)} {\small\textbf{.752}} & \textbf{(3)} {\small\textbf{.560}} \\
        \midrule
        Claim (text) & 16.029 & 2.688 & 0.0613 & 0.0049 \\
        Claim (PD) & - & 0.994 & - & 0.0296 \\
        Claim (TE) & - & - & 0.0899 & 0.0441 \\
        \midrule
        Evidence (text) & 5.279 & 0.4434 & 0.0007 & 0.001 \\
        Evidence (PD) & - & 0.3213 & - & 0.008 \\
        Evidence (TE) & - & - & 0.005 & 0.008 \\
        \bottomrule
    \end{tabular}
    }
  \caption{Predicted label distribution and absolute attribution strengths. Note that strengths for evidence are for a single evidence document.}
  \label{table:table}
  \end{subfigure}
\caption{Illustration of BiLSTM ($glob$) attribution strengths for an example taken from MultiFC.}
\label{fig:attribution}
\end{figure}

\paragraph{Importance of time in final prediction}

While we have shown that temporal reasoning strongly influences relevance estimations and label scores per evidence document, we now measure how much the time-aware fact-checking models rely on temporal information for their final veracity predictions. For this, we attribute the prediction of the models to the input using integrated gradients \citep{pmlr-v70-sundararajan17a}. This attribution technique measures the attribution strength of text and time features on the final prediction. We focus on the base BiLSTM model and its best-performing temporal variants. 
Given the high dimensionality of text and time embeddings, the attribution strengths across all dimensions are summed to obtain an total attribution value for claim, evidence, and time ($c_t$ and $e_{i,t}$). 
Figure \ref{fig:attribution} illustrates the attribution values of a single data entry and presents the ranking of evidence text and time according to their attribution strength. The models typically attributed the prediction to both the claim and evidence, with a stronger emphasis on the collected evidence than on the claim. However, when time information was introduced, the attribution strength of claim and evidence texts strongly decreased, especially when evidence was grounded at the content level (CL/DL+CL). This indicates that time indeed influences model prediction. 

Interestingly, the attribution ranking of temporal information was found to be distinct from that of the content, as demonstrated by the example in Figure \ref{fig:attribution}. The publication dates that are closer to that of the claim obtain higher attribution strength than those far from the claim. 
In line with this, statistical correlation testing between $e_{i,t}$ and label scores $q_i$ - where each label score in $q_i$ for $e_{i,t}$ in the same bucket is compared to the label score in $q_i$ for $e_{i,t}$ in different buckets - show that evidence contained within the same time bucket tend to prefer the same prediction labels as their label rankings strongly 
correlate ($\rho$ = 0.7). We can thus conclude that time influences both interim and final prediction.

\section{Conclusion}

Grounding claims and associated evidence documents on a shared timeline and implicitly reasoning over their temporal relations noticeably improves the verification performance of automated fact-checking models. 
Time plays a dual role in this process, serving both as a source of information for verifying claims, as well as influencing the evaluation of the relevance and supporting or refuting nature of evidence documents.
Further research may look into integrating temporal reasoning in claim detection and evidence retrieval processes or implementing even more sophisticated temporal reasoning during claim verification by examining the temporality of events discussed in a claim and their relation to the evidence.

\section*{Limitations}

The limitations of this work mainly originate from the data and the use of pretrained models for grounding claims and evidence documents in time. Since the evidence documents were retrieved after the claim had been fact-checked by giving the claim verbatim to a search engine and selecting the first ten search results, their quality and relevance to the claim is not ensured. As a result, evidence-based fact-checking models risk relying on spurious signals in the evidence documents for predicting a claim's veracity. Moreover, the evidence documents are presented as short snippets which only reflect small parts of the full Web documents. This not only affects content representation, but it also limits temporal information extraction since many time expressions may have been omitted from the shortened text. Regarding temporal information extraction and normalisation, we had to rely on pretrained models to obtain temporal representations of claims and its associated evidence documents. This not only introduces noise in the input data, but also requires time-expensive preprocessing. 

\section*{Ethics Statement}

Automated fact-checking technology aims to assist people in distinguishing between verified and unverified content in professional contexts and during their daily information consumption. Nevertheless, the fact-checking models constructed in this paper - like all fact-checking models - should be deployed with caution and its predictions should never be taken as final without further human evaluation. Computational predictions are anything but flawless, and incorrect predictions may unjustly discredit the person or group who uttered the fact-checked statement(s).  

\section*{Acknowledgements}
This work was realised with the collaboration of the European Commission Joint Research Centre under the Collaborative Doctoral Partnership Agreement No 35332. The scientific output expressed does not imply a policy position of the European Commission. Neither the European Commission nor any person acting on behalf of the Commission is responsible for the use which might be made of this publication.
The research leading to this paper also received funding from 
the European Research Council (ERC) under Grant Agreement No. 788506.
The resources and services used in this work were provided by the VSC (Flemish Supercomputer Center), funded by the Research Foundation - Flanders (FWO) and the Flemish Government.

\bibliography{anthology,custom}
\bibliographystyle{acl_natbib}

\appendix

\section{Time Buckets} \label{app:timebuckets}

Table \ref{tab:bpub} presents an overview of time buckets $b_{pub}$ with their interval bounds used for document-level grounding, while Table \ref{tab:btext} presents time buckets $b_{text}$ with their interval bounds used for content-level grounding.

\begin{table*}
\centering
\small
\begin{tabular}{|l|l|c|}
\hline
\multicolumn{3}{|l|}{\normalsize \textbf{Overview of time buckets for document-level grounding and reasoning: $b_{pub}$}}\\
\hline
 \textbf{Start} & \textbf{End} & \textbf{Number of evidence documents}\\
 \hline
 $\infty$ days before the claim & 1596 days before the claim & 8536 \\
 \hline 
  1596 days before the claim & 741 days before the claim& 8547 \\
  \hline
  740 days before the claim & 361 days before the claim & 8528\\
  \hline
  360 days before the claim & 146 days before the claim & 8517 \\
  \hline 
  145 days before the claim & 35 days before the claim & 8626\\
  \hline 
  34 days before the claim & 4 days before the claim & 8962\\
  \hline
  3 days before the claim & 1 day before the claim & 7549 \\
  \hline
  on the same day as the claim & on the same day as the claim & 8963\\
  \hline
  1 day after the claim & 4 days after the claim & 8735\\
  \hline
  5 days after the claim & 24 days after the claim & 8548\\
  \hline
  25 days after the claim & 85 days after the claim & 8345 \\
  \hline
  86 days after the claim & 187 days after the claim & 8534 \\
  \hline
  188 days after the claim & 325 days after the claim & 8551 \\
  \hline
  326 days after the claim & 498 days after the claim & 8515 \\
  \hline
  499 days after the claim & 736 days after the claim & 8502 \\
  \hline 
  737 days after the claim & 1061 days after the claim & 8533\\
  \hline
  1062 days after the claim & 1436 days after the claim & 8529 \\
  \hline
  1437 dagen na de claim & 1997 days after the claim & 8537 \\
  \hline 
  1998 days after the claim & 2605 days after the claim & 8531 \\
  \hline
  2606 days after the claim & $\infty$ days after the claim & 8522\\
  \hline
\end{tabular}
\caption{\normalsize Overview of time buckets $b_{pub}$ with their interval bounds.}
\label{tab:bpub}
\end{table*}

\begin{table*}
\centering
\small
\begin{tabular}{|l|l|c|}
\hline
\multicolumn{3}{|l|}{\normalsize \textbf{Overview of time buckets for content-level grounding and reasoning: $b_{text}$}}\\
\hline
 \textbf{Start} & \textbf{End} & \textbf{Number of evidence documents}\\
 \hline
 $\infty$ days before the claim & 18172 days before the claim & 12853 \\
 \hline 
  18171 days before the claim & 6295 days before the claim& 12851 \\
  \hline
  6294 days before the claim & 2928 days before the claim & 12856\\
  \hline
  2927 days before the claim & 1678 days before the claim & 12862 \\
  \hline 
  1677 days before the claim & 989 days before the claim & 12855\\
  \hline 
  988 days before the claim & 569 days before the claim & 12863\\
  \hline
  568 days before the claim & 323 days before the claim & 12833 \\
  \hline
  322 days before the claim & 145 days before the claim & 12935 \\
  \hline
  144 days before the claim & 42 days before the claim & 12771 \\
  \hline
  41 days before the claim & 6 days before the claim & 13191 \\
  \hline
  5 days before the claim & 1 day before the claim & 13269 \\
  \hline
  on the same day as the claim & on the same day as the claim & 22966\\
  \hline
  1 day after the claim & 8 days after the claim & 15135\\
  \hline
  9 days after the claim & 42 days after the claim & 12665\\
  \hline
  43 days after the claim & 124 days after the claim & 12832 \\
  \hline
  125 days after the claim & 241 days after the claim & 12739 \\
  \hline
  242 days after the claim & 378 days after the claim & 12888 \\
  \hline
  379 days after the claim & 581 days after the claim & 12828 \\
  \hline
  582 days after the claim & 834 days after the claim & 12852 \\
  \hline 
  835 days after the claim & 1178 days after the claim & 12862\\
  \hline
  1179 days after the claim & 1582 days after the claim & 12834 \\
  \hline
  1583 days after the claim & 2134 days after the claim & 12848 \\
  \hline 
  2135 days after the claim & 2734 days after the claim & 12848 \\
  \hline
  2735 days after the claim & $\infty$ days after the claim & 12842\\
  \hline
\end{tabular}
\caption{\normalsize Overview of time buckets $b_{text}$ with their interval bounds.}
\label{tab:btext}
\end{table*}

\section{Reproducibility Settings}
\label{sec:reproducibility}

This section contains settings for reproducing the experiments in this paper.

\paragraph{Computing infrastructure}

The BiLSTM models were trained on a Skylake processor type with one compute node, 9 cores per node, one GPU (GPU partition of Skylake) and 5 GB memory per core. The DistilRoBERTa models were trained on a Cascadelake processor type with one compute node with 4 cores per node, one GPU and 5 GB memory per core.

\paragraph{Average runtime}

Preprocessing, i.e., extraction of timex annotations via Heideltime, open information extraction (where before this a correction of uppercase characters is done via Named Entity Recognition), and construction of the dataset where claims and evidence are already put into buckets and the predicates and timexes are marked in the text of all the data took approximately 150 hours. Training a BiLSTM model for each domain took on average 45 hours, while a DistilRoBERTa model took 72 hours.

\paragraph{Number of model parameters}

BiLSTM: 16,129,125 learnable parameters per model; DistilRoBERTa: 82,933,601 learnable parameters per model.

\paragraph{Number of training and evaluation runs} 

Without parameterisation by $\alpha$, $\beta$, and $\gamma$: 150 epochs pretraining, 100 epochs fine-tuning (both BiLSTM and DistilRoBERTa).With parameterisation: 600 epochs pretraining, 300 epochs fine-tuning (BiLSTM); 800 epochs pretraining, 300 epochs fine-tuning (DistilRoBERTa).

\begin{figure*}[ht]
    \centering
    \includegraphics[scale=0.4]{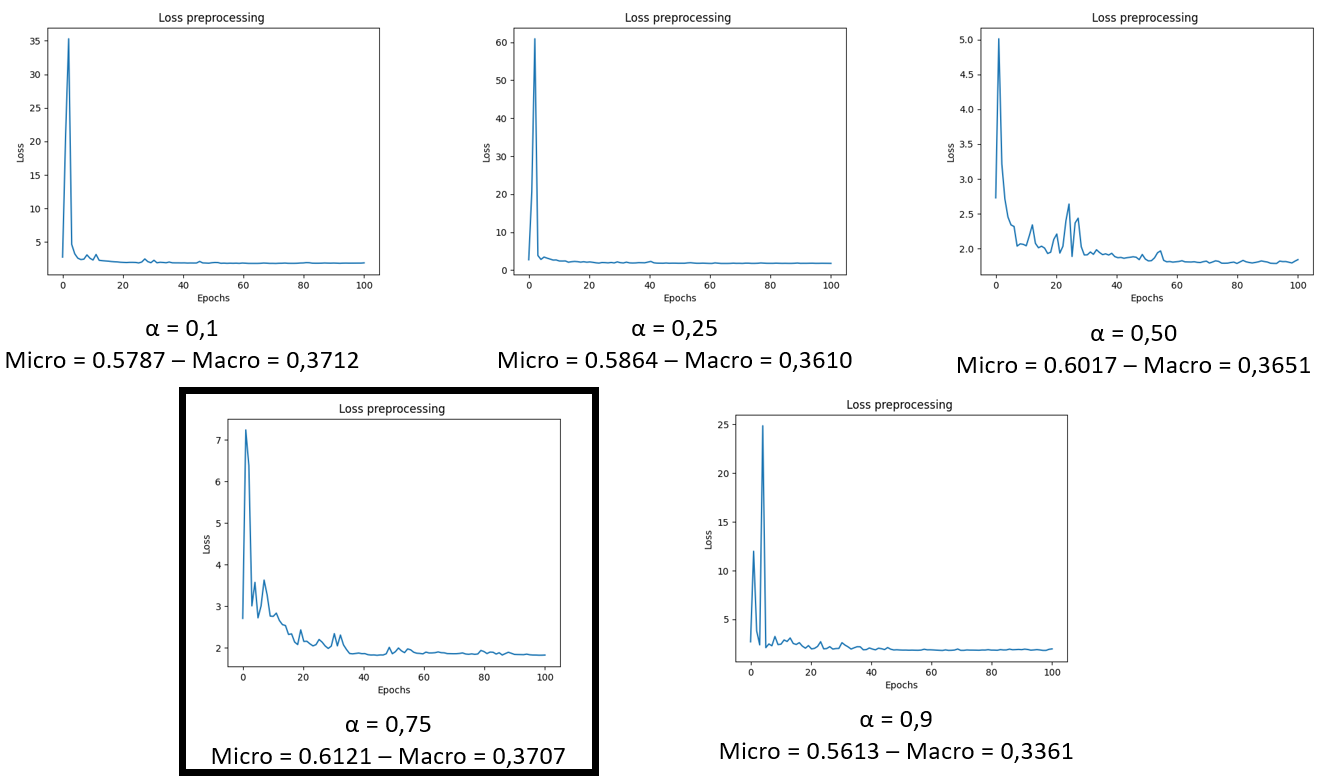}
    \caption{Tuning $\alpha$ for DistilRoBERTa (DL$_{glob}$) based on the prediction performance on the validation set (metrics: Micro/Macro F1) and the validation loss.}
    \label{fig:hypertuning}
\end{figure*}

\paragraph{Hyperparameter bounds} 

We \textit{manually} tested following combinations for $\alpha$ when integrating the time attribution vectors at the global level for document-level (DL) or content-level reasoning: $\alpha \in \{0.10,0.25,0.50,0.75,0.90\}$. Final $\alpha$-values:
BiLSTM (DL$_{glob}$): $\alpha=0.10$; BiLSTM (CL$_{glob}$): $\alpha=0.25$; DistilRoBERTa (DL$_{glob}$): $\alpha=0.75$ (see Figure \ref{fig:hypertuning}); DistilRoBERTa; DistilRoBERTa (CL$_{glob}$): $\alpha=0.10$. We tested following combinations for $\gamma$ when integrating the time attribution vectors at the local level for content-level reasoning (CL): $\gamma \in \{0.10,0.25,0.50,0.75,0.90\}$. Final $\gamma$-values:
BiLSTM (CL$_{loc}$): $\gamma=0.50$; DistilRoBERTa (CL$_{loc}$): $\gamma=0.10$.
We tested following combinations for $\alpha$ and $\beta$ when grounding the time attribution vectors at both the document and content level (DL+CL): $[(\alpha=0.20,\beta=0.20),(\alpha=0.20,\beta=0.35),(\alpha=0.20,\beta=0.40),(\alpha=0.20,\beta=0.55),(\alpha=0.20,\beta=0.60),(\alpha=\frac{1}{3},\beta=\frac{1}{3}),(\alpha=0.35,\beta=0.20),(\alpha=0.35,\beta=0.55),(\alpha=0.40,\beta=0.20),(\alpha=0.40,\beta=0.40),(\alpha=0.55,\beta=0.20),(\alpha=0.55,\beta=0.35),(\alpha=0.60,\beta=0.20)]$. Final $\alpha$- and $\beta$-values: BiLSTM (DL+CL$_{glob}$): $(\alpha=0.20,\beta=0.40)$; DistilRoBERTa (DL+CL$_{glob}$): $(\alpha=0.20,\beta=0.20)$. We performed a hyperparameter search trial of 100 epochs pretraining for each combination of hyperparameters. The criterions used to select the final hyperparameter values are the prediction performance (Micro/Macro F1) on the validation loss and the evolution of the validation loss (visualised on a plot, see Figure \ref{fig:hypertuning}).

\paragraph{Other parameters tested} 

\begin{itemize}
    \item Without linear scheduler;
    \item With linear scheduler with warm up;
    \item With linear learning scheduler;
    \item Learning rates: 0.001, 0.005, 0.0002 (only for RMSprop), 0.0001, 0.00001 (for pretraining and fine-tuning);
    \item Adam, RMSProp (Only BiLSTM), AdamW (only DistilRoBERTa);
    \item With weight decay: 0.001, 0.0001;
    \item Without weight decay.
\end{itemize}

\end{document}